# Multiple Feature Fusion-based Video Face Tracking for IoT Big Data

Tianping Li, Zhifeng Liu, Jianping Qiao*

Tianping Li, Key Laboratory of Medical Physics and Image Processing in Shandong Province, Shandong Provincial Key Laboratory for Distributed Computer Software Novel Technology, Shandong Normal University, School of Physics and Electronics, Jinan, Shandong, China(e-mail: sdsdltp@163.com).

Zhifeng Liu, School of Physics and Electronics, Shandong Normal University, Jinan, Shandong, China (e-mail: liuzhifengne@163.com).

Jianping Qiao, Corresponding author, Associate Professor at School of Physics and Electronics, Shandong Normal University, Jinan, China (e-mail: jpqiao@sdu.edu.cn ).

**Abstract:** With the advancement of IoT and artificial intelligence technologies, and the need for rapid application growth in fields such as security entrance control and financial business trade, facial information processing has become an important means for achieving identity authentication and information security. In this paper, we propose a multi-feature fusion algorithm based on integral histograms and a real-time update tracking particle filtering module. First, edge and colour features are extracted, weighting methods are used to weight the colour histogram and edge features to describe facial features, and fusion of colour and edge features is made adaptive by using fusion coefficients to improve face tracking reliability. Then, the integral histogram is integrated into the particle filtering algorithm to simplify the calculation steps of complex particles. Finally, the tracking window size is adjusted in real time according to the change in the average distance from the particle centre to the edge of the current model and the initial model to reduce the drift problem and achieve stable tracking with significant changes in the target dimension. The results show that the algorithm improves video tracking accuracy, simplifies particle operation complexity, improves the speed, and has good anti-interference ability and robustness.

## 1 Introduction

Face tracking technology is an important computer vision research area and is widely used in the fields of the Internet of Things and artificial intelligence technologies. Face tracking is the process of predicting the motion information in subsequent frames based on the motion information in the initial frame of a face to determine the face trajectory and its morphological changes. Traditional face tracking mainly matches for a single feature, extracting a single colour feature, edge feature, texture feature or motion information[1][2], with low robustness. With the rapid development of the Internet of Things in the 21st century, target detection and target tracking are widely used in the Internet of Things and artificial intelligence technologies[3], intelligent surveillance, Intel-Link traffic, human-machine interfaces, etc., and play a crucial role in electronic intelligence, security, military and other fields. Therefore, the current research on multiple features has far-reaching significance, and facial recognition has the advantages of simple acquisition, difficult to copy, and stealing complex; however, video data are generally outdoors, and video images are relatively small and of poor quality, so video facial recognition technology requires high real time, and video face tracking is still an extremely challenging problem in the field of computer vision[4]. How to quickly and accurately detect faces and implement real-time video face tracking algorithms has become an important research process for face information processing.

In this paper, the video face tracking[5] method is improved and refined, in which colour features and edge features are adaptively fused to improve the face tracking reliability, the integral histogram method improves efficiency, the size of the tracking window is adjusted in real time according to the average distance from the particle centre to the edge of the current model and the initial model and the change in the tracking target[6], the tracking module is updated to reduce the drift problem, and finally, the face video tracking is tested using a dataset to test the method. The experimental data show that the method accuracy is further improved, the particle operations complexity is simplified, and it has good stability and robustness under the influence of video with object occlusion, lighting changes, and similar backgrounds and finally achieves accurate real-time human face tracking.

The rest of this paper is organized as follows. In Section 2, two important particle filtering algorithm processes for video face tracking[7] are briefly described. In Section 3, the improved resampling-based particle filtering algorithm proposed in this paper, as well as the methods for building the image integration histogram and fusing multiple features are described in detail. Section 4 presents the face tracking system, compares the experimental results qualitatively and quantitatively. Finally, the paper is summarized, and future tracking algorithms and future work are envisioned.

**2 Particle filtering algorithm**

Particle filtering (PF) is characterized by high accuracy because it is not limited by noise and system models. Problems that cannot be solved by traditional analysis methods can be solved with the help of particle filtering simulation, and in recent years, particle filtering algorithms have been very successful in the field of target tracking technology[8]. PF is based on Bayesian inference and importance sampling theory as the basic framework.[9] The Bayesian inference process uses a collection of particles to represent the posterior probability distribution of the random state of the target and estimate the state of the nonlinear system. The essence of importance sampling is to select larger weights for particles with high trust and thus, determine the probability of being a target based on the specific distribution of particle weights.

In this paper, we study the tracking of a single face, and the algorithm obtains the maximum likelihood observation through the ideal likelihood observation peak distribution and uses the given system model to estimate the present moment state from the previous moment state to obtain the closest estimate of the true target state, enhance the observation effect and the adaptive capability of the target tracking, and predict the posterior probability $p(x_t | y_{1:t})$ of the target state at time t. The particle filtering steps are divided into four processes: initial state, target prediction, state correction and resampling.

Initial state: A large number of particles is used to simulate the uniformity of the particle distribution[10].

Prediction phase: Particle filtering generates a large number of particles based on the probability distribution of $x_{t-1}$, and the prior probability density of the state is predicted using the state transfer equation and the control volume to make a preliminary estimate of the system at the next moment through the existing prior knowledge. The prediction equation is shown in Eq. (1).

$$p(x_t | y_{1:t-1}) = \int p(x_t | x_{t-1}) p(x_{t-1} | y_{1:t-1}) dx_{t-1} \qquad (1)$$

Correction stage: The posterior probability density is introduced using the similarity between the possible and real target states and the observation equation, and the closer it is to the real state, the larger the particle weight is, otherwise, the smaller the weight is that is assigned. The update equation is shown in Eq. (2).

$$p(x_t \mid y_{1:t}) = \frac{p(y_t \mid x_t) p(x_t \mid y_{1:t-1})}{p(y_t \mid y_{1:t-1})} \tag{2}$$

Using a recursive approach to reduce the complexity when calculating the weights, the signal is processed by sequential importance sampling, and the distribution condition of the importance function at moment (t-1) is used as the importance function at moment t. The particle weight $\{w_t^1, w_t^2, \ldots, w_t^N\}$ recursive form can be expressed as:

$$\begin{aligned} w_t^i &\propto \frac{p(x_{0:t}^i \mid y_t)}{q(x_{0:t}^i \mid y_t)} = \frac{p(y_t \mid x_t^i) p(x_t^i \mid x_{t-1}^i) p(x_{0:t-1}^i \mid y_{t-1})}{q(x_t^i \mid x_{0:t-1}^i, y_t) q(x_{0:t-1}^i \mid y_{t-1})} \\ &= w_{t-1}^i p(y_t \mid x_t^i) \end{aligned} \tag{3}$$

Resampling: After sequential importance sampling, in the iterative process the diversity of particles is lost due to the difference between the importance function and the posterior distribution. To alleviate the degradation problem, based on adjusting the importance function, the particles are filtered according to the particle weights, and the importance weight of the particles is replicated instead of the number of low-weight particles for propagation to obtain the required true state. In the particle filtering algorithm, it is not necessary to consider resampling at each iteration, but the effective number of particles $N_{eff}$ is cited to determine the degradation size of the algorithm. If $N_{eff} < N_{th}$, the resampling algorithm is used, $N_{th}$ is the threshold value, which is generally taken as 2N/3, and the value is usually approximated by $\hat{N}_{eff}$.

$$\hat{N}_{eff} = \frac{N}{\sum_{i=1}^{N} \omega(x_{0:t})^2} \tag{4}$$

The posteriori density can be expressed as:

$$p(x_t \mid y_{1:t}) \approx \sum_{i=1}^{N} \widetilde{w}_t^i \delta(x_t - x_t^i) \tag{5}$$

Finally, the resampled particles are brought into the state transfer equation again for the prediction process. The general framework structure of the particle filtering algorithm is shown in Figure 1.

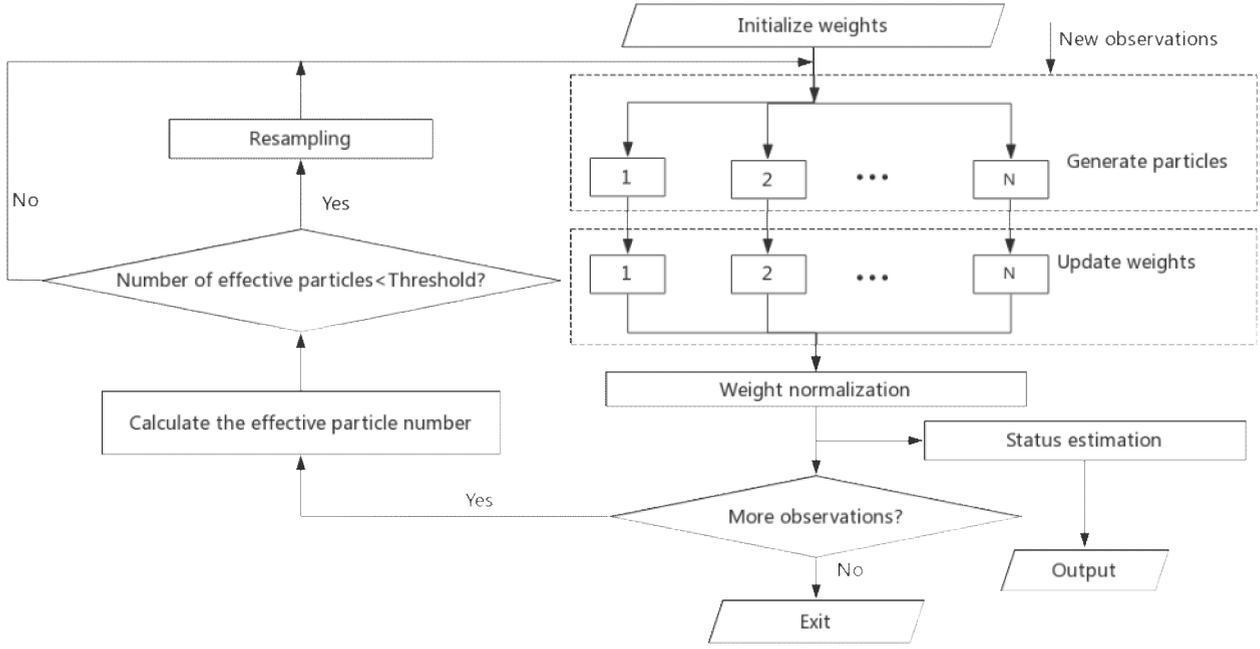

Figure 1. General framework structure of the particle filtering algorithm

## 3 Improved particle filtering algorithm

The traditional resampling algorithm only resamples the particles according to certain particle weight rules, without considering the particles distributed in the high likelihood region, and the small weight samples fail[11], leading to particle degradation and new large estimation errors. In response to the negative effects created by resampling techniques, an improved resampling algorithm is proposed to replace the traditional particle resampling algorithm, improve the particle filtering algorithm performance, adaptively adjust the particles distributed in the high likelihood region according to the accuracy factor value reflecting the measurement noise[12], increase existing knowledge and the same region of likelihood, and improve the algorithm robustness.

### 3.1 Specific steps for resampling

Pre-processing stage: Choose time $\{x_k^i\}_{i=1}^N$ to obtain the required particle set, obtain the particle pair according to the formula $w_k^{i+1} > w_k^i$, reset the formula, reweight the new particles to ensure that the particles meet the needed particle weights, and through this process, obtain the new particle pair.

Particle classification: Set the weight $w_t$ and the threshold $w_h$ to satisfy the relation $(0 < w_l < w_h)$ by which the particles are divided into two major parts.

The range class A particle values: $\{(x\,k\,i, w_k^j); w_k^j \leq w_l, w_k^j \geq w_h, i=1,2,\ldots,N\}$. The range class B particle values: $\{(x\,k\,i, w_k^j); w_l < w_k^j < w_h, i=1,2,\ldots,N\}$. Gravity particles and small weight particles form class A particles, and stable medium weight particles form class B particles. The sum of two particles is the sum of all particles.

Handling particles: Resampling class A particles using linear equations:

$$x_n = x_s + KL(x_a - x_s) \qquad (6)$$

$$L = \left[\frac{1}{N_{AW}}\right]^{\frac{1}{m}} \tag{7}$$

where $x_n$ represents the new particle, and the heavy particle with more times and repetitions is selected; $x_s$ is the small and medium weight particle; $K$ is the step coefficient, and the value of $K$ is determined according to the actual needs during the experiment. The new weight particles are generated by $(x_a - x_s)$ and compared with the original particle weight. If the new particle weight is small, we reduce $L$ by 1/2 and generate a new particle according to the new $L$.

**3.2 Simulation experiments and data analysis**

Simulation experiments verify the effectiveness of the improved resampling technique for state estimation and tracking nonlinear systems using the TRPF algorithm and IRPF algorithm, respectively. The experiments use the unary nonstationary growth model (UNGM), and the state model and observation model equations of the simulation object are as follows:

$$x_n = 0.5x_{n-1} + \frac{25x_{n-1}}{1+x_{n-1}^2} + 8\cos[1.2(n-1)] + u_n \tag{8}$$

$$y_n = \frac{1}{20}x_n^2 + v_n \tag{9}$$

where $u_n$ and $v_n$ are zero-mean Gaussian noise, the simulation takes the number of particles as 100, the step size n=50, the coefficient K=0.4, and the measurement noise variance is 1 for 50 iterations.

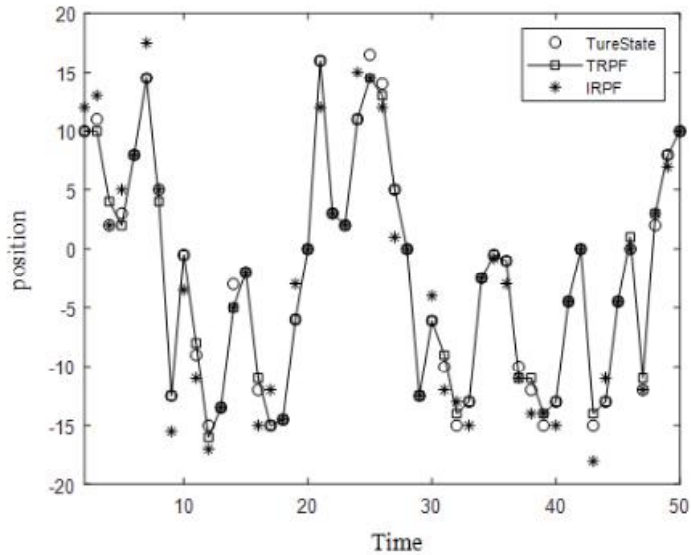

Figure 2. IRPF algorithm state estimation

The specific state estimation curves are shown in Figure 2. Under the same simulation environment, the TRPF algorithm with traditional resampling has a poor estimation effect because of the small number of particles, the concentration of the likelihood function is higher than that of the state transfer density function, and the particles are severely depleted, which leads to a poor match between the state estimation and the real state. However, the state estimation of the improved particle filtering algorithm IRPF proposed in this paper can match the real state well with many effective samples and high estimation accuracy, which can effectively suppress sample degradation and ensure particle diversity compared with the TRPF algorithm.

To better compare the performance of the improved resampling particle filtering algorithm in terms of state estimation, the mean square error (RMSE) is used as a measure, and the RMSE is calculated as:

$$RMSE = \left[\frac{1}{T}\sum_{K=1}^{T}(x_k - \hat{x}_k)\right]^{\frac{1}{2}} \qquad (10)$$

where $T$ represents the time step of an experiment, and $x_k$ and $\hat{x}_k$ represent the true and estimated values at moment $K$. The smaller the RMSE value, the better the effect.

Figure 3 shows the comparison of RMSE results for 50 independent experimental simulations.

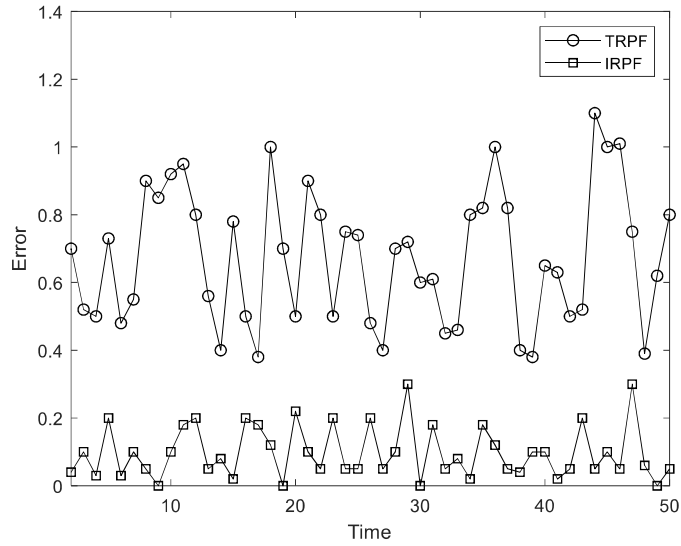

**Figure 3. Comparison of RMSE results between the TRPF algorithm and IRPF algorithm**

A comparative plot of the RMSE results shows that the TRPF algorithm using traditional resampling has a significantly higher root mean square error and greater fluctuations than the IRPF algorithm in this paper, which uses a shorter state estimation time. In summary, the IRPF algorithm performs much better than the TRPF algorithm with good robustness.

**4 Face tracking system**

**4.1 Power model**

The tracking algorithm is based on a moving target model, and the autoregressive process model AR(p) has been widely used to create such a dynamic model. The AR(p) model is abbreviated as:

$$x_t = \phi_0 + \phi_1 x_{t-1} + \phi_2 x_{t-2} + \ldots + \phi_p x_{t-p} + \varepsilon_t \qquad (11)$$

The target displacement, noise, velocity and acceleration are all attributes of the moving target. Therefore, a second-order regression process is used to describe the motion law. The AR(2) model is

$$x_t = \phi_1 x_{t-1} + \phi_2 x_{t-2} + \varepsilon_t \qquad (12)$$

where $\{x_t, t = 0, \pm 1, \pm 2, \ldots\}$ is the time series, $\phi_1$ and $\phi_2$ are the drift coefficient matrices[13], $\{\varepsilon_t\}$ is the white noise series, and $\phi(B)x_t = \varepsilon_t$, where $\phi(B) = 1 - \phi_1 B - \phi_2 B^2$. Let the linear transfer function be $\psi(B)$. The smooth solution of the second-order regression process is:

$$x_t = \psi(B)\varepsilon_t \qquad (13)$$

These parameters can be obtained empirically or by training video sequences, and the weight coefficients $x_t$ of the function $\psi(B)$ can be found[14]. The feasibility of the model is guaranteed, and whether the particle propagation is reasonable needs to be verified in the particle update process.

**4.2 Observation model**

In particle filtering, observation is feature-based. This section focuses on facial feature extraction by combining colour features and edge features to track[15] the target face.

**4.2.1 Observation model based on colour features**

The colour histogram shows the proportion of different colours in an image[16]. Colour information has the features of translation, rotation and computational simplicity compared to geometric features and is currently widely used in face detection and tracking problems. In this paper, human skin colour is used as a coarse detection and localization method for human faces, and the colour distribution is described more robustly by using HSV(Hue, Saturation, Value) colour space modelling and computing histograms using only face elements. For skin colour detection, the skin colour points are judged according to a threshold value. The colour histogram is shown in Figure 4(b). The effect of illumination variation is reduced to some extent.

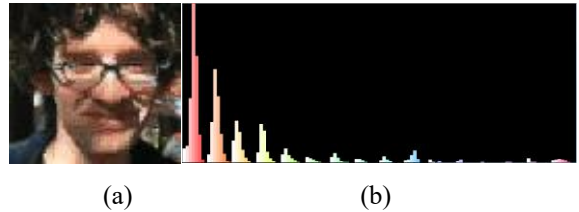

(a)          (b)

**Figure 4. Colour histogram**

The target region weighted colour histogram is constructed as:

$$P_l(n) = C_l \sum_{i=1}^{N} K_E\left(\frac{x_i - x_0}{a}\right) \delta[h(x_i) - n] \qquad (14)$$

$$C_l = \frac{1}{\sum_{i=1}^{N} K_E\left(\frac{x_i - x_0}{a}\right)} \qquad (15)$$

$C_l$ is a normalization constant, N is the total number of pixels in the target region, and $K_E(.)$ is an Epanechnikov kernel contour of radius $n \in [1, M]$, indicating that the closer a pixel is to the centre of the target, the more likely it is that it belongs to the target, and the Epanechnikov kernel is described as follows:

$$K_E(x) = \begin{cases} C(1-\|x\|^2), & \|x\| < 1 \\ 0, & \|x\| \geq 1 \end{cases} \quad (16)$$

$\delta_a$ is the Kronecker delta function, function $h(x_i)$ maps the pixel positions to the corresponding histogram face elements, and n is the histogram segment number index value.

When the colour distributions of the reference target template $q_l(n)$ and the candidate target template $p_l(n)$ are calculated, the Bhattacharyya distance $d_l$ can be used to measure the similarity of the two distributions as follows.

$$d_l = \sqrt{1 - \rho_l[p_l(n), q_l(n)]} \quad (17)$$

where $\rho_l[p_l(n), q_l(n)]$ is the Bhattacharyya dispersion coefficient:

$$\rho_l[p_l(n), q_l(n)] = \sum_{n=1}^{M} \sqrt{p_l(n) q_l(n)} \quad (18)$$

The Bhattacharyya coefficient measures the correlation, M is the total number of colour-weighted histogram bins, and $x_0$ is the centre of the observed region. The face colour observation likelihood probability density function[17] is defined as:

$$p(y_l | x) = \frac{1}{\sqrt{2\pi}\sigma_l} e^{\left(-\frac{d_l^2}{2\sigma_l^2}\right)} \quad (19)$$

where $\sigma_l$ is the Gaussian variance, which is taken as 0.2 in the experiment. The larger the value of Eq. (19), the greater the colour similarity between the candidate target and the face template, and the greater the possibility that it is a real face target.

**4.2.2 Observation model based on edge features**

Another key information cue of the face is the edge feature[18], which is characterized by insensitivity to illumination changes and similar face colour backgrounds, and the edge direction histogram reflects the edge and texture features of the face.

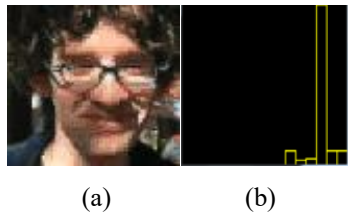

(a)        (b)

**Figure 5. Histogram of edge directions**

The histogram of edge directions is shown in Figure 5(b). First, the original image is greyed out where the edge points are extracted by the operator operation, and the shape features of the image are obtained using the statistical edge point direction histogram. In this paper, the Sobel operator is used to detect the edge contours, the amplitude of the gradient $g$ is calculated by Eq. (20), and the gradient direction $\theta$ is calculated by Eq. (21).

$$g = \sqrt{g_x^2 + g_y^2} \quad (20)$$

$$\theta = \arctan \frac{g_y}{g_x} \quad (21)$$

where $g_x$ and $g_y$ are the horizontal and vertical gradients of the image, respectively.

$$\rho_m[p_m(n), q_m(n)] = \sum_{n=1}^{M} \sqrt{p_m(n) q_m(n)} \quad (22)$$

The face edge likelihood function is defined as:

$$p(y_m | x) = \frac{1}{\sqrt{2\pi}\sigma_m} e^{\left(-\frac{1-\rho_m}{2\sigma_m^2}\right)} \quad (23)$$

where $\sigma_m$ is the Gaussian variance, which is taken as 0.3 in the test.

**4.3 Strategy of this paper**

Based on the improved resampling-based particle filtering, a multi-feature fusion algorithm based on an integral histogram and a real-time update tracking particle filter module is proposed for video face tracking, which not only reduces the algorithm running time and drift problem but also improves the tracking accuracy and robustness to achieve a real-time tracking effect considering the long computation time.

**4.3.1 Adaptive multi-feature fusion**

Video face tracking finds the region in the current frame image that has the most similar template features to the tracked target[19]. Nummiaro et al. extracted colour to verify the tracking effect, and the experiment showed that the single colour information is insensitive to object occlusion but very sensitive to illumination. In this paper, we combine the edge features, which are insensitive to illumination changes and similar face colour background and propose a strategy of multi-feature information fusion based on improved resampling particle filtering to calculate the particle weights. First edge features and colour features are extracted in the video dataset, and a weighting method is used to weight the colour and edge features histogram to describe the facial features. Then, the fusion coefficient is used to adaptively fuse the colour features and edge features[20]. Adaptive particle filtering removes the excess number of particles to improve the computational speed and the reliability of the face observation model to make up for the respective deficiencies of colour features and edge features, which achieves complementary information between features and achieves video target tracking in complex backgrounds. The observation model $p(y_t | x_t)$ extracts colour features and edge features. The whole face likelihood function is defined as:

$$p(y_t | x_t) = \theta_l p(y_l | x) + \theta_m p(y_m | x) \quad (24)$$

where $\theta_l + \theta_m = 1$, the weight $\theta_m$ is large when the edge information is reliable in tracking; otherwise, vice versa. $p(y_l | x_t)$ is the likelihood function of the colour feature, and $p(y_m | x)$ is the likelihood function of the edge feature.

The particle weights are updated as:

$$w_t^i \propto w_{t-1}^i p(y_t | x_t^i) = w_t^i [\theta_l p(y_l | x) + \theta_m p(y_m | x)] \quad (25)$$

**4.3.2 The integral histogram of the image**

In the particle filter tracking algorithm, when there are many particles, the computational time consumed will be greater. To solve this problem, the integral histogram is integrated into the particle filtering algorithm, which only needs to add and subtract the integral histogram of the four vertices of the rectangular region where the particles are located, which replaces the inefficient statistical work of the ordinary algorithm, simplifies the calculation steps of complex particles, solves the problem of long histogram calculation time caused by too many particles, and improves the operational efficiency.

To verify the effectiveness of the integral histogram, we analyse the effect of the ordinary histogram and the integral histogram on the computational speed with different numbers of particles.

| Number of particles | Time/s | |
|---|---|---|
| | Normal histogram | Integral histogram |
| 20 | 0.028987 | 0.050296 |
| 50 | 0.085672 | 0.054322 |
| 100 | 0.148562 | 0.058970 |
| 500 | 0.765326 | 0.063952 |

**Table 1. Effect of different particle numbers on the calculation**

As seen in Table 1, the computational time of the ordinary histogram is temporarily shorter when the number of particles is low, and as the number of particles increases, i.e., the area of the region where the particles are located increases, the integral histogram used in this paper converges faster and the computational time is shorter. The strategy of using the integral histogram in this paper is more robust than the ordinary histogram because the number of particles selected in this paper is higher.

**4.3.3 Calibration model**

Since the face state continues to change during the tracking process, current face information is reflected in real time by updating the template, and frequent updating of the template; this causes the real target in the template to be replaced by the background and the drifting phenomenon [21] will occur, which eventually leads to tracking failure. To mitigate drift, in this paper, we propose a template correction technique to correct the template matching result according to the change in the average distance from the particle centre to the edge of the current model and the initial model, adjust the size of the tracking window in real time, and stably track face targets with significant dimensional changes. The template adaptive correction method used in this paper is as follows:

$$H_{new} = \frac{1}{\tau} H_{old} + (1 - \frac{1}{\tau}) H_{current} \tag{26}$$

where $H_{new}$ is the new reference histogram, $H_{old}$ is the initial reference histogram, and $H_{current}$ is the current reference histogram. $\tau$ is the constant $(0 \leq 1/\tau \leq 1)$, and the smaller $\tau$ is, the more effective the histogram update is. The algorithm can also update the tracking module in real time according to the changes in the tracking target, thus improving the accuracy and precision of real-time tracking.

**4.4 Experimental results**

To verify the tracking effectiveness of the method in this paper, three sets of more complex experimental videos, including object occlusion, illumination changes, and similar backgrounds, are used to track two different algorithms, namely, the colour-only-based tracking algorithm and the tracking algorithm in this paper. We manually select a rectangular window as the matching template in the window and set N=100. The experimental platform is based on Visual Studio 2010 opencv2.4.8 and uses Euclidean distance to measure the target tracking results. The dataset is used for the visual tracker benchmark, and the complete benchmark contains 100 sequences from the recent literature. The video sequences in this paper are obtained from the references, and the information of the video sequences is shown in Table 2.

| Test | Sequence | Frame size | Sequence characteristics | Total frames | Total frames in this paper |
|------|----------|------------|--------------------------|--------------|----------------------------|
| Test1 | FaceOcc2 | 480×360 | Object occlusion | 812 | 310 |
| Test2 | Trellis | 480×360 | Illumination variation | 569 | 182 |
| Test3 | David | 480×360 | Similar background | 770 | 119 |

**Table 2. Video sequences**

### 4.4.1 Test 1

Test 1 investigates the effect of face tracking under object occlusion.

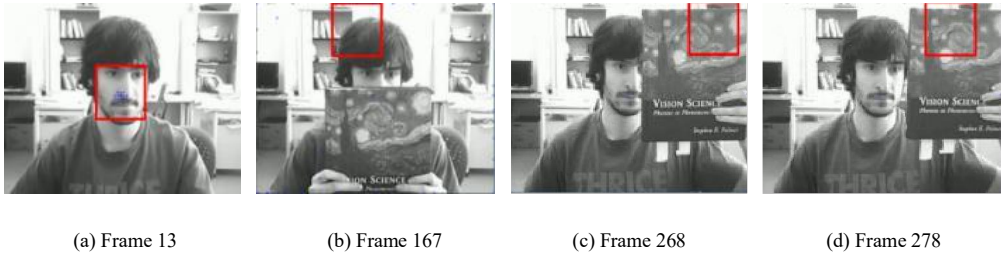

(a) Frame 13   (b) Frame 167   (c) Frame 268   (d) Frame 278

**Figure 6. Tracking results using colour information alone when objects are occluded**

Figure 6 shows the tracking results based on colour features only when the face is occluded by other objects. When unobscured, the face position can be accurately located. However, when the face is partially occluded by the book, the weights change drastically, the face region information is reduced, and the tracking algorithm based on colour features alone tends to lose the face in frame 167 of Figure 6(b). When the book moves to change the occlusion position, the localization ability is completely lost. When the occluded object is completely removed, it still fails to locate the real position of the face in time, and the tracking effect is poor.

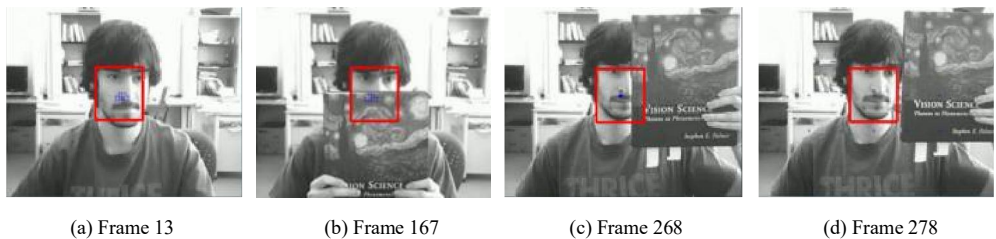

(a) Frame 13   (b) Frame 167   (c) Frame 268   (d) Frame 278

**Figure 7. Tracking results of the algorithm in this paper when the object is obscured**

The tracking results of the method used in this paper are shown in Figure 7. In frames 167 and 268 in Figure 7(b)(c), the tracking box can accurately locate the face region in the cases of object occlusion and when the occlusion moves, and the algorithm continues to track the face accurately after the occlusion is withdrawn[22].

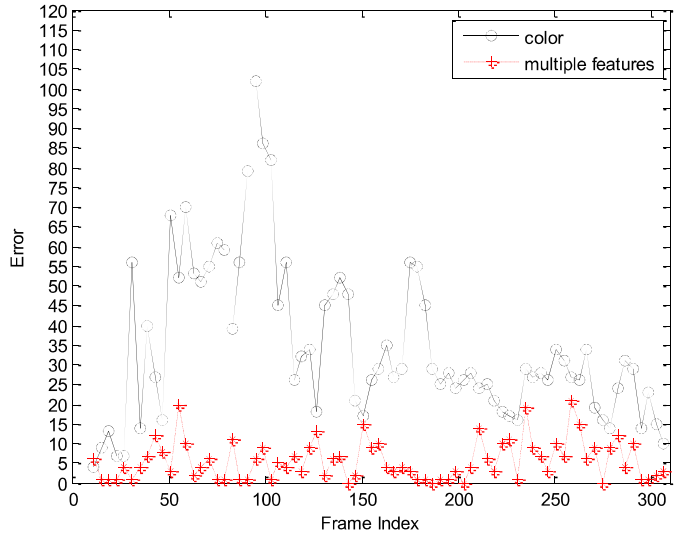

**Figure 8. Comparison of RMSE for single features and multiple features based on object occlusion**

The comparison curves of the root mean square error RMSE between single and multi-features are given in Figure 8. The tracking error of the algorithm in this paper is significantly lower than the tracking error based on colour features only. This is mainly due to the proposed fusion scheme, which enhances the tracking reliability.

#### 4.4.2 Test 2

Test 2 investigates the effect of face tracking under different illumination conditions.

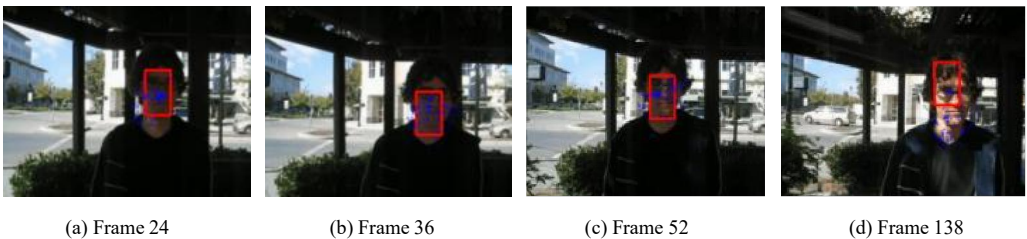

(a) Frame 24    (b) Frame 36    (c) Frame 52    (d) Frame 138

**Figure 9. Tracking results using colour information alone when lighting changes**

Tracking based on colour features is better under different illumination only before the light changes, and when the changes are significant, the tracking results deviate from the real face position. In frames 36 and 138 in Figure 9(b)(d), when the experiment is based on colour features only, tracking fails. This is because the tracking algorithm based on colour features only is very sensitive to illumination changes.

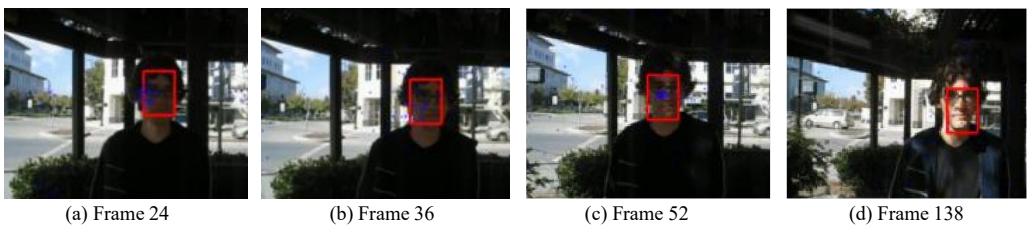

(a) Frame 24    (b) Frame 36    (c) Frame 52    (d) Frame 138

**Figure 10. Tracking results of the algorithm in this paper when the illumination changes**

Figure 10 shows that the adaptive multi-feature fusion strategy proposed in this paper can track faces well even when the illumination changes because the combined edge features are insensitive to the change in face surface colour, which compensates for the effect of illumination changes on the tracking results and reduces error.

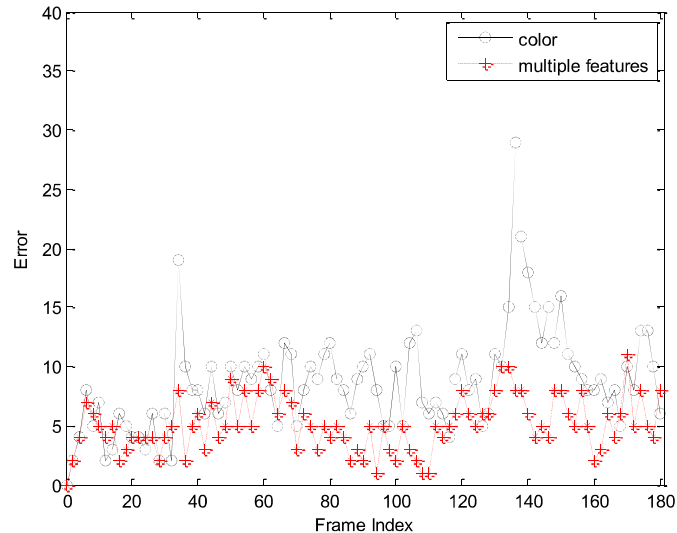

**Figure 11. Comparison of RMSE of a single feature and multiple features based on light variation**

Figure 11 shows our quantitative analysis of the mean square error tracking results. The results show that the adaptive fusion strategy and the real-time update tracking particle filter module in this paper have a smaller overall error value with fluctuation, are insensitive to the change in a light colour and have a better tracking effect compared with the algorithm that simply extracts single features.

### 4.4.3 Test 3

Test 3 investigates the effect of face tracking in backgrounds similar to the face.

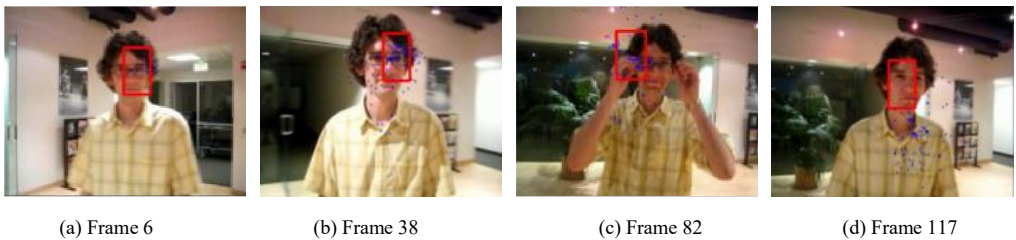

(a) Frame 6    (b) Frame 38    (c) Frame 82    (d) Frame 117

**Figure 12. Tracking results using colour information alone with backgrounds similar to the face**

Figure 12 shows the tracking results based on colour features only when a background similar to the face is present. If only colour features are used for particle filtering, the algorithm is insensitive to facial expressions and pose changes, but when an object similar to the face appears in the video, for example, in frame 82 of Figure 12(c), the face is disturbed by the hand, which causes the algorithm to misjudge, and the rectangular box cannot locate the correct face position and tracks the hand[23]. The face can be tracked only when the similar background disappears.

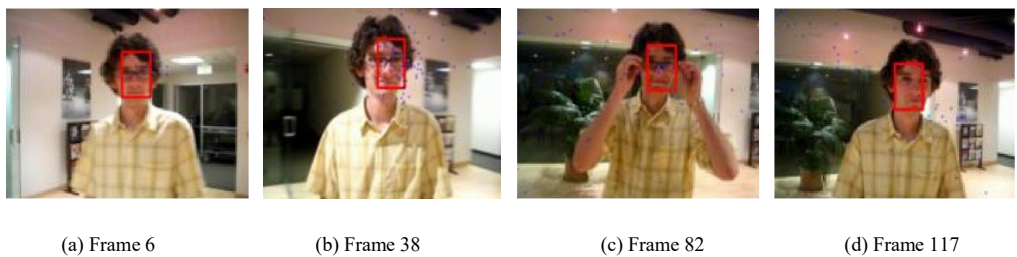

| (a) Frame 6 | (b) Frame 38 | (c) Frame 82 | (d) Frame 117 |

**Figure 13. Tracking results of this paper's algorithm in similar backgrounds**

Figure 13 shows the tracking effect of the method used in this paper, which can track faces stably even with a similar background. It can also be seen in Figure 8 and Figure 11 that our improved algorithm has very small error and better anti-interference ability and robustness.

| Test | Running time/s | | |
| --- | --- | --- | --- |
| | Colour-based | Colour-based and edge-based | Our algorithm |
| Test1 | 100.9 | 136.8 | 109.9 |
| Test2 | 55.8 | 84.5 | 61.6 |
| Test3 | 15.9 | 30.6 | 21.2 |

**Table 3. Comparison of experimental running times of different algorithms**

A comparison of the experimental runtimes of the different algorithms is shown in Table 3. The running speed based on colour features alone is the fastest, but the tracking accuracy is poor. The above video experiment results of the adaptive multi-feature fusion tracking algorithm based on colour and edge features can accurately track the face target while ensuring a shorter runtime. Considering the long computational time, this paper uses an integral histogram and correction model to compute the facial features based on adaptive multi-feature fusion[24], which not only reduces the algorithm running time and drift problem but also improves the tracking accuracy and robustness to achieve a real-time tracking effect.

## 5 Conclusion

In the process of acquiring facial feature information, face information is easily affected by factors such as object occlusion, lighting changes, and similar backgrounds, for target tracking in complex backgrounds, it is often difficult to achieve robust tracking with a single feature algorithm. In this paper, we use the fused colour and edge information to describe the observation information of the target, combine the fused observation model to the particle filtering algorithm based on improved resampling, First, edge and colour features are extracted, weighting methods are used to weight the colour histogram and edge features to describe facial features, and fusion of colour and edge features is made adaptive by using fusion coefficients to improve face tracking reliability. And introduce the integral histogram and real-time update tracking particle filtering module to make the algorithm converge faster, have shorter computational time, and track more stably. Finally, the tracking window size is adjusted in real time according to the change in the average distance from the particle centre to the edge of the current model and the initial model to reduce the drift problem and achieve stable tracking with significant changes in the target dimension. Experiments show that the method in this paper can precisely locate the face position under conditions of object occlusion, lighting change, similar colour background, etc., and save time compared with extracting a single feature. For the face occlusion problem, the tracking algorithm used in this paper is most effective by updating the template.

However, the proposed algorithm has some limitations. For example, hardware-assisted methods need to be considered, the initial template of the face should be well defined or the target model should be anchored to the first frame. With the rapid development of science and technology, target tracking has become an inevitable trend. How to detect and accurately track the face in the video is a key part of information processing, and the study of target tracking under real natural conditions is still of great value. For the target tracking based on particle filtering algorithm, many researches have been done and certain research results and effects have been achieved. With the rapid development of deep learning, the target detection and tracking model based on convolutional neural network has richer features and stronger feature expression ability. The target tracking algorithm faces many challenges, and in the future, the target tracking algorithm will be further studied, and a series of improvements based on deep learning algorithm and on deep learning algorithm will research faster and more robust tracking algorithms.

**Author Contributions:** original draft preparation, Z.L.; writing review, T.L.; communication, J.Q. All authors have read and agreed to the published version of the manuscript.

**Funding:** Key Laboratory of Medical Physics and Image Processing in Shandong Province

**Conflicts of Interest:** The authors declare that they have no conflict of interest in this work, that we have no business or joint interests that present a conflict of interest in connection with the work submitted, no financial or personal relationships with other persons or organizations that might unduly influence our work, and no professional or other personal interests of any nature or kind in any products, services, or companies.